\documentclass[11pt]{article}

\usepackage[final]{acl}
\usepackage{times}
\usepackage{latexsym}
\usepackage[T1]{fontenc}
\usepackage[utf8]{inputenc}
\usepackage{graphicx}

\graphicspath{{../figures/}{figures/}{paper/figures/}}

\newcommand{\ours}{\textnormal{AgenticRAG-FP}}
\newcommand{\dr}{\textnormal{Doctor-RAG}}
\newcommand{\pa}{\textnormal{Propagation-Aware}}
\newcommand{\lj}{\textnormal{LLM-Judge}}
\newcommand{\rb}{\textnormal{Rule}}
\newcommand{\sr}{\textnormal{Suf-Regen}}

\title{When Failures Propagate: Causal Failure Attribution\\
in Agentic Retrieval-Augmented Generation}
\author{Lauren Pothuru \\
  Cornell University \\
  Anote AI \\
  \texttt{lsp75@cornell.edu} \And
  Spurthi Setty \\
  Stevens Institute of Technology
  \And
  Natan Vidra \\
  Anote AI}
\date{}

\begin{document}
\maketitle

\begin{abstract}
Agentic retrieval-augmented generation (RAG) interleaves retrieval, reasoning,
and answer generation across multiple hops. A retrieval error at hop 1 can
surface only as a wrong answer at hop 3, while later retrieval can also repair
the trajectory. This paper introduces \ours, an interventional benchmark for
causal failure attribution in agentic RAG. The benchmark injects a certified
fault at a specified hop, re-executes the downstream trajectory, and evaluates
diagnosers against the known intervention site. Its central question is whether
a post-hoc trace still identifies the injected hop after the suffix changes. In
the dense Claude Haiku 4.5 structural sweep on three-hop MuSiQue
questions, coverage-based diagnosis is 0.91 at hop 1 and 0.00 at hops 2 and
3 ($n{=}43,36,21$ failed trajectories), but it predicts hop 1 in 96\% of
failures, an earliest-hop rule; at hops 2 and 3, an LLM judge and frozen-hop
repair localize only 0.25 to 0.48 of failures. In larger
content-corruption runs, depth-2 exact-hop accuracy is 0.00 for coverage,
0.49 for an LLM judge, 0.59 for frozen-hop repair, and 0.14 for suffix
regeneration ($n{=}51$); the judge and frozen-hop repair do not significantly
differ (exact McNemar $p{=}0.33$).
\end{abstract}

\section{Introduction}

Retrieval-augmented generation (RAG) conditions generation on retrieved
evidence rather than only model parameters \citep{lewis2020rag}. Agentic RAG
extends this process across several decisions: a language model selects a
sub-query, observes evidence, and decides whether to search again or answer,
as in ReAct-style reasoning and acting \citep{yao2022react}. This enables
multi-hop question answering, but it also separates a visible end-to-end error
from its cause.

Consider a three-hop trajectory. If the evidence at hop 1 is irrelevant, the
agent can form a drifted sub-query at hop 2, retrieve evidence on the wrong
premise, and produce an ungrounded final answer. An evaluator that inspects
only the final trace may identify a late low-coverage hop or label the failure
as answer generation, while missing the earliest causal fault. A different
agent may retrieve compensating evidence at a later hop and answer correctly.
Standard answer accuracy, retrieval recall, and final-answer correctness
conflate an injected fault, its propagation, and its recovery.

Existing diagnostic methods can inspect traces or attempt local repair, but
post-hoc traces do not provide a certified intervention-site label. This leaves two
evaluation questions unresolved: can a diagnoser recover a known intervention
hop after the failure has propagated, and when do counterfactual repair probes
supply information missing from the final trace? Answering either question
requires an intervention that allows the agent to react and retains the
intervention label.

\ours{} provides this setting. It corrupts a trace prefix at hop $h$, runs the
remaining trajectory with the same agent, and scores diagnoses against the
certified intervention. The paper centers on how propagation changes the
information available for exact-hop attribution. Its three contributions are:
\begin{enumerate}
  \item An interventional benchmark that supports structural retrieval faults
  and certified content corruption. The latter changes a known fact span in
  otherwise relevant evidence and records the span pair, enabling deterministic
  absorbed, resisted, and derailed outcome labels without an LLM judge.
  \item A depth-conditioned analysis of post-hoc signal. In the dense MuSiQue
  sweep, coverage-based attribution falls from 0.91 at hop 1 to 0.00 at hops 2
  and 3, but predicts hop 1 in 96\% of failures, an earliest-hop rule. After
  suffix re-execution, no evaluated diagnoser exceeds 0.48 exact-hop accuracy
  at structural hops 2 and 3.
  \item A larger comparison of counterfactual probes on content
  faults. At depth 2 ($n{=}51$), coverage, judge, frozen-hop, and
  suffix-regeneration accuracy are 0.00, 0.49, 0.59, and 0.14; the judge
  and frozen-hop repair do not significantly differ (exact McNemar $p{=}0.33$).
  Depth-3 active-probe cells are budget-censored, not zero-accuracy findings.
\end{enumerate}

\section{Related Work}

\paragraph{RAG and active retrieval.}
RAG combines a parametric generator with a non-parametric retrieval module
\citep{lewis2020rag}. Active and reflective variants decide when to retrieve
during generation \citep{jiang2023flare,asai2023selfrag}. These approaches
aim to improve answer quality by making retrieval responsive to uncertainty or
the evolving generation. That same responsiveness creates the attribution
problem studied here: a change to an early retrieval can alter the questions
the agent asks later, rather than simply changing one fixed evidence set.

\paragraph{Agentic RAG and diagnosis.}
ReAct-style agents interleave reasoning traces with external actions
\citep{yao2022react}. Their trajectories can be inspected, localized, and
repaired, but inspection alone cannot establish whether a suspicious hop
caused the final answer. Doctor-RAG uses coverage-gated localization and
prefix reuse for diagnosis and repair \citep{jiao2026doctorrag}; \dr{}
instantiates its coverage-localization idea as a post-hoc baseline. The present
evaluation differs in its target: it asks whether a diagnoser recovers a
certified injected cause after an agent has generated a new suffix. This
requires observing both the intervention and the resulting trajectory.

\paragraph{Multi-hop benchmarks.}
HotpotQA \citep{yang2018hotpotqa} and MuSiQue \citep{trivedi2021musique}
provide multi-hop questions with supporting facts. FRAMES
\citep{krishna2024frames} evaluates factuality, retrieval, and reasoning, and
CRAG \citep{yang2024crag} includes dynamic, long-tail, and false-premise
questions. These datasets provide question and corpus substrates, but their
standard answer labels do not identify the hop that caused a failed trajectory.
The benchmark in this paper supplies that missing causal label through a
controlled intervention.

\paragraph{Causal evaluation.}
The intervention perspective follows causal evaluation
\citep{pearl2009causality}: the fault and its hop are set before the resulting
trajectory is observed. The object of intervention is not a model parameter or
an internal activation. It is the trajectory-level context available to a
retrieving agent. This framing makes recovery informative rather than
inconvenient, because a correct answer after intervention is evidence that the
agent did not propagate the fault to its output.

\section{Task and Trace Model}
\label{sec:task}

Each example is a question-answer pair $(q,y^*)$ together with a retrieval
corpus $\mathcal{C}$. An agent produces a trace
\[
\tau=\bigl(q,\{(q_h,D_h)\}_{h=1}^{H},A,y,c\bigr),
\]
where $q_h$ is the sub-query issued at hop $h$, $D_h\subseteq\mathcal{C}$ is
the retrieved evidence, $A$ is the final answer, $y$ is the reference answer,
and $c$ is total token cost. The trace records the information available at
each decision and is sufficient to resume execution from an observed prefix.

\paragraph{Failure stages.}
Each hop has a stage
\[
s\in\{\mathrm{retrieval},\;\mathrm{tool},\;\mathrm{answer},\;\mathrm{none}\}.
\]
Retrieval failures include empty or irrelevant evidence, query drift,
false-premise evidence, stale evidence, and a wrong fact in otherwise relevant
documents. Tool failures include missing actions and premature termination.
Answer failures include empty answers, incorrect answers, and hallucinations
that are unsupported by the retrieved evidence. The \emph{none} stage denotes
a trace without an identified failure at that hop. A diagnosis record contains
a predicted stage $\hat{s}$, predicted hop $\hat{h}$, propagation flag,
severity estimate, and intervention-site description.

\paragraph{Identifiability.}
A diagnoser $d$ is identifiable at depth $h$ when its exact-hop accuracy is
measured on failed injected traces whose certified injection depth is $h$:
$\hat{h}=h$. This is stricter than detecting that the final answer is wrong or
naming a broad failure stage. The target is the \emph{injected hop}, not a
semantic cause inferred from the final trace: after suffix regeneration, a
later hop can carry the corrupted dependency even though the certified
intervention occurred earlier. Appendix~\ref{app:example} gives a worked
three-hop instance.

The benchmark treats the selected failure type, stage, and hop as certified
labels. Certification matters because a late low-coverage hop can be a cause,
a downstream effect, or a benign consequence of an earlier intervention.
Without a known intervention, these possibilities are observationally similar.
The live comparison fixes the intervention before execution and then evaluates
the diagnosis only after the agent has had an opportunity to propagate or
repair its effect.

\section{Interventional Benchmark}
\label{sec:benchmark}

\paragraph{Resumable live intervention.}
Given an executed hop prefix, the agent resumes at the next hop and continues
to generate sub-queries, retrieve evidence, and decide when to answer. A live
intervention replaces evidence or a sub-query at hop $h$, then resumes the
agent at hop $h+1$. The downstream suffix is therefore an agent response to
the corrupted context, not a post-hoc trace edit. This distinction separates
the causal effect of the intervention from the behavior of a static edited
trace, which cannot reveal whether an agent would propagate or recover from a
fault.

For every live intervention, the evaluation retains the intended failure
family and selected hop as the target label. The intervention can be applied to
retrieval evidence, a sub-query, or the decision to terminate. Structural
faults test whether a visible retrieval anomaly remains localizable after the
suffix changes. Content faults test the harder case in which the document still
looks relevant and only its factual content is wrong. The same trace model
therefore supports both readily observable and semantically hidden failures.

\begin{table}[t]
\centering
\small
\begin{tabular}{p{0.26\linewidth}p{0.58\linewidth}}
\hline
Intervention & Effect at the selected hop \\
\hline
Empty retrieval & Replace retrieved evidence with $\emptyset$. \\
Irrelevant documents & Replace evidence with off-topic documents. \\
Query drift & Substitute a drifted sub-query, then retrieve. \\
False premise & Add a confidently wrong factual claim to evidence. \\
Stale evidence & Add outdated temporal evidence. \\
Content corruption & Change a certified fact span in otherwise relevant evidence. \\
Early termination & Force an answer from the preceding evidence. \\
\hline
\end{tabular}
\caption{Interventions used by \ours. In every live intervention, the
downstream suffix is generated after the selected-hop corruption.}
\label{tab:interventions}
\end{table}

\paragraph{Intervention semantics.}
Empty and irrelevant retrieval remove answer support or replace it with
off-topic evidence. Query drift changes the search intent before retrieval.
False-premise and stale-evidence interventions preserve retrieved material but
introduce a specific misleading claim. Early termination changes the available
trajectory by requiring an answer before a later retrieval could occur. These
families span missing evidence, misleading evidence, and altered agent
decisions. They also differ in how visible they are to a post-hoc inspection,
which makes them useful tests of whether a diagnoser is recognizing a local
symptom or identifying an intervention after propagation.

All interventions preserve the selected-hop target while allowing the suffix
to change. An answer error following an intervention is thus not assumed to be
caused by an edited final answer. It is an observed outcome of the agent's
reaction to the modified prefix. This construction also permits a recovered
trajectory to remain part of the evaluation: recovery is evidence about the
agent's resilience, while failed trajectories are the subset on which the
intervention site can be localized.

\paragraph{Certified content corruption.}
Structural interventions replace whole documents or queries, so they can be
visible in the trace. Content corruption instead changes one fact in retrieved
documents while preserving topical relevance. This models an error such as a
stale fact, a corrupted document, or an upstream extraction error without
making the selected retrieval obviously off-topic.

The changed span is selected by priority. An \emph{answer fact} is selected
when the gold answer occurs in the retrieved evidence and can be replaced by a
certified wrong value. Otherwise, a \emph{bridge entity} is selected when it
occurs both in the hop evidence and in a later sub-query, but not in the
original question. This identifies information that the agent carried forward.
A \emph{salient entity} or number supplies a fallback when neither condition
holds. Numeric spans receive deterministic perturbations; entity spans receive
in-domain distractors with no token overlap with the original span.

Corruption is seeded per trace and hop, so the selected replacement is
reproducible. Samples with no certifiable span are skipped and counted rather
than receiving an unverified change. The recorded original and corrupted spans
certify the intervention and make deterministic answer-level evaluation
possible. They also make it possible to distinguish copying a wrong fact from
producing another unrelated error.

\paragraph{Diagnosers.}
The main comparisons use a coverage-based localizer (\dr{}), a trace-reading
LLM judge (\lj{}), frozen-hop counterfactual repair (\pa{}), and
suffix-regeneration repair (\sr{}). \dr{} scores gold-final-answer token
coverage per hop and returns the earliest below threshold; intermediate hops
typically lack the final answer. A rule-based diagnoser (\rb{}) sees only the
final trace, attributes the observed failure stage, and uses an earliest-empty-hop
heuristic; it is reported only as an appendix reference. The frozen-hop probe repairs one candidate
hop by re-running retrieval for its original sub-query against the clean
per-question corpus, falling back to the original question if that retrieval
is empty. \pa{} holds all other hops fixed and forces an answer; \sr{} keeps
the repaired prefix and resumes the agent to generate the suffix. Thus the two
probes differ in their treatment of downstream evidence. \dr{}, \lj{}, and
both active probes receive the reference answer. They are therefore
oracle-assisted benchmark instruments rather than deployable diagnosers. Active
probes test candidates earliest first; the structural sweep uses a four-probe
budget, so at most the first four candidate hops are tested.

\section{Metrics}
\label{sec:metrics}

\paragraph{Attribution identifiability.}
For injected traces that remain incorrect after suffix re-execution, exact-hop
accuracy for diagnoser $d$ is
\[
\mathrm{Acc}_d(h)=\frac{1}{|\mathcal{F}_h|}
\sum_{\tau_i\in\mathcal{F}_h}1[\hat h_{d,i}=h_i],
\]
where $\mathcal{F}_h$ is the set of failed injected traces at depth $h$.
Recovered trajectories are excluded because there is no final failure to
attribute. Accuracy is accompanied by 95\% percentile bootstrap confidence
intervals using $B{=}1{,}000$ resamples clustered by base question, because
multiple interventions can share a base question. Cells with fewer than 10
failed traces are descriptive rather than comparative. The content depth-3
post-hoc estimates have $n{=}35$; active depth-3 cells are
budget-censored because a two-probe budget never tests hop 3.

The denominator is deliberately the set of traces that remain failed after
live re-execution. Including recovered trajectories as incorrect diagnoses
would conflate intervention-site localization with the separate question of whether a
fault reached the answer. Conversely, reporting only answer accuracy would
discard the information carried by a certified intervention. This conditional
metric makes the two outcomes visible: recovery measures resilience, and
attribution measures localization among failures that actually propagated.
The coverage localizer, LLM judge, and both active probes receive the gold
answer; these are oracle-assisted instruments rather than deployable methods.

\paragraph{Localization beyond exact hop.}
Exact-hop accuracy is intentionally demanding when faults are causally
entangled across hops. Stage accuracy asks whether $\hat{s}=s$.
Hop-tolerance accuracy accepts $|\hat{h}-h|\leq1$. Ancestor-hit rate counts a
prediction when $\hat{h}\leq h$ and $\hat{s}=s$, giving partial credit when a
predicted earlier hop is a causal ancestor of the injected fault. Mean absolute
hop error summarizes the distance between $\hat{h}$ and $h$. These metrics
separate broad failure recognition from localization of the intervention.
They are secondary diagnostics and are not used for the central empirical
claim.

\paragraph{Counterfactual recovery.}
Recovery is the fraction of live interventions whose corrupted trajectory
still produces a correct answer:
\[
\mathrm{Recovery}(h)=
\Pr\bigl[\mathrm{correct}(A,y)\mid do(f,h)\bigr].
\]
It is a separate robustness outcome. High recovery means that an intervention
did not propagate into an end-to-end failure, not that a diagnoser localized it
successfully.

\paragraph{Deterministic generation outcomes.}
For content faults, the certified span pair supports a judge-free answer-level
classification:
\[
\begin{array}{ll}
\mathrm{absorbed} & A\ \mbox{contains changed-span tokens},\\
\mathrm{resisted} & \mbox{if } \mathrm{correct}(A,y),\\
\mathrm{derailed} & \mbox{otherwise.}
\end{array}
\]
Correctness takes precedence when an answer contains both the gold and
corrupted values. Query contamination records whether the corrupted span
appears in a later sub-query, providing an observable channel for propagation.
The outcome labels answer a different question from exact-hop attribution.
They characterize the effect of a known content change on the generated answer
even when the resulting trajectory is correct or has too little evidence for a
localization comparison. Because the labels are computed from the certified
span pair, a value can be checked consistently across backbones without
introducing a second model's interpretation of whether the error was copied.

\paragraph{Cost per correct diagnosis.}
For token-spending diagnosers, total diagnosis tokens are divided by the
number of correct localizations. This distinguishes a method's localization
quality from deployability: an accurate method can still be unsuitable for
online diagnosis if each correct localization requires disproportionate
oracle-assisted re-execution or judging.

\section{Experimental Setup}
\label{sec:setup}

\paragraph{Datasets and corpora.}
HotpotQA and MuSiQue are anchor multi-hop question-answering datasets with
annotated supporting facts. FRAMES supplies variable-depth RAG questions with
Wikipedia passage corpora, and CRAG contributes dynamic, long-tail, and
false-premise questions. Headline FRAMES conditions use fetched and cached
Wikipedia passage text. Conditions that use only link titles as a stand-in
corpus are excluded from headline comparisons and flagged separately because
that substitute changes retrieval fidelity.

\paragraph{Backbones and retrieval.}
The main structural result uses Claude Haiku 4.5
with dense retrieval on three-hop MuSiQue examples and a four-probe budget.
The repository also contains BM25, GPT-4o-mini, and local-model conditions,
but incomplete or low-failure-count cells are not used for the main comparison.
Static controls use BM25 and token-overlap retrieval; their results are
retained as appendix controls, not as the central causal experiment.

\paragraph{Structural interventions and depth eligibility.}
The structural experiment injects empty retrieval, irrelevant retrieval,
false-premise evidence, and stale evidence at hops 1, 2, and 3. For each
sample and requested depth, all four intervention types apply to the same base
trace, then the downstream suffix is re-executed. A depth-specific case is
eligible only when the base trace answered correctly and the actual injected
hop equals the requested depth. The primary structural summary uses the
available records that support these checks. The matrix is not yet a complete
factorial sweep across every backbone, retriever, and depth, so the results are
reported as evidence about the observed post-hoc signal rather than as a
universal method ranking.

The failed scoring sets differ across depths: they contain 22, 18, and 10 base
questions, and their pairwise base-question Jaccards are 0.29 (hops 1--2),
0.14 (1--3), and 0.22 (2--3). Intervention-family mixtures also shift
modestly: empty/irrelevant/false-premise/stale counts are 12/14/11/6,
13/9/6/8, and 8/5/4/4. Thus depth rows condition on different
surviving-failure subpopulations. They show post-hoc signal under each realized
failure set, not a within-question depth curve; a matched-question analysis is
future work.

\paragraph{Content-corruption conditions.}
Content-corruption runs\footnote{These runs use larger samples than the review
version of this paper (failed $n{=}44/18/3$ there) and supersede its
content-fault estimates.} use GPT-4o-mini and Claude Haiku 4.5 on HotpotQA and
MuSiQue with BM25. Each MuSiQue condition starts from 140
examples, uses depths 1--3, and requires declared depth at least 3; each
HotpotQA condition starts from 120 examples and uses depths 1--2 by design.
Traces from one model family are judged by the other. The pooled failed counts
are 91, 51, and 35, with depth 3 MuSiQue-only. The active probes use a
two-probe budget and earliest-first search, so they never test hop 3; their
depth-3 entries are budget-censored rather than 0.00 findings.

\paragraph{Evaluation units.}
For structural conditions, a requested depth and intervention family define an
evaluation unit over eligible base traces. For content conditions, the span
selection strategy additionally determines whether a fault changes an answer
fact, bridge entity, or salient fallback. All diagnosers receive the same
post-intervention trace for a unit. Active probes then create their own
counterfactual repairs from that trace, so their additional evidence is part of
the diagnoser rather than a change to the benchmark label.

\section{Results}
\label{sec:results}

\subsection{Structural Attribution Across Depths}
\label{sec:depth-results}

The structural sweep shows coverage zeros at later hops after suffix
re-execution, but its earliest-hop rule means that these alone do not establish
loss of post-hoc signal. Table~\ref{tab:dense-claude} reports all three depths
for dense Claude Haiku 4.5 on MuSiQue. The per-cell $n$
values are failed injected trajectories eligible for exact-hop scoring; each
estimate includes its nonparametric bootstrap 95\% interval.

\begin{table*}[t]
\centering
\scriptsize
\begin{tabular}{lccc}
\hline
Diagnoser & Hop 1 ($n{=}43$) & Hop 2 ($n{=}36$) & Hop 3 ($n{=}21$) \\
\hline
\dr{} & 0.91 [0.80,0.98]; 1.00; 0.09 & 0.00 [0.00,0.00]; 1.00; 1.00 & 0.00 [0.00,0.00]; 0.00; 2.00 \\
\lj{} & 0.26 [0.10,0.42]; 0.67; 1.07 & 0.25 [0.13,0.38]; 0.81; 0.94 & 0.43 [0.19,0.71]; 0.48; 1.43 \\
\pa{} & 0.51 [0.36,0.67]; 0.86; 0.67 & 0.25 [0.11,0.38]; 0.86; 0.89 & 0.48 [0.24,0.75]; 0.52; 1.05 \\
Always hop 1 & 1.00 [1.00,1.00]; 1.00; 0.00 & 0.00 [0.00,0.00]; 1.00; 1.00 & 0.00 [0.00,0.00]; 0.00; 2.00 \\
\hline
\end{tabular}
\caption{Exact-hop attribution for the dense Claude Haiku 4.5 MuSiQue
sweep. Each cell is exact-hop accuracy [base-question-clustered 95\% CI];
within-$\pm1$-hop accuracy; MAE. $n$ is failed injected trajectories.}
\label{tab:dense-claude}
\end{table*}

At hop 1, coverage identifies the deliberately visible structural fault. At
later hops, coverage is 0.00 in both cells, while the judge and frozen-hop
probe retain partial signal. \dr{} predicts hop 1 for 100\% of failed cases at
depths 2 and 3 (90.7\% at depth 1), so its hop-1 success does not exceed the
always-hop-1 prior; the informative cells are its deep zeros. The judge selects
hop 3 for 14/43 (0.33) depth-1 cases, a late-hop tendency. \pa{} is
non-monotonic (0.51, 0.25, 0.48), so active accuracy is not a simple depth
curve. The result does not support a general method ordering.

\subsection{Counterfactual Probe Performance}
\label{sec:counterfactual-results}

The content-fault study examines how two counterfactual scopes behave when the
evidence remains topically relevant. Table~\ref{tab:content-acc} pools the four
conditions. At depth 2, the four diagnosers are co-located for
comparison: coverage is 0.00, the judge is 0.49, frozen-hop repair is 0.59,
and suffix regeneration is 0.14 ($n{=}51$). The judge predicts hop 2 in 0.20
of depth-1 failures and 0.03 of depth-3 failures, versus 0.49 at depth 2, so
its content-fault predictions track the injected hop rather than a fixed
middle-position prior. \pa{} exceeds the judge in three of four
depth-2 conditions and ties it in the fourth; the GPT-4o-mini--HotpotQA cell
has only $n{=}6$. Appendix~\ref{app:content-by-condition} gives the breakdown.

\begin{table}[t]
\centering
\scriptsize
\resizebox{\linewidth}{!}{%
\begin{tabular}{lccc}
\hline
Diagnoser & Depth 1 ($n{=}91$) & Depth 2 ($n{=}51$) & Depth 3 ($n{=}35$) \\
\hline
\dr{} (coverage) & 1.00 [1.00,1.00] & 0.00 [0.00,0.00] & 0.00 [0.00,0.00] \\
\lj{} (cross-family) & 0.43 [0.33,0.53] & 0.49 [0.35,0.63] & 0.57 [0.41,0.73] \\
\pa{} & 0.88 [0.80,0.95] & 0.59 [0.45,0.72] & --$^\ast$ \\
\sr{} & 0.92 [0.86,0.97] & 0.14 [0.04,0.24] & --$^\ast$ \\
\hline
\end{tabular}
}
\caption{Exact-hop attribution on content faults, pooled over
GPT-4o-mini and Claude Haiku 4.5 on HotpotQA and MuSiQue with BM25.
Brackets are base-question-clustered 95\% CIs; depth 3 is MuSiQue-only.
$^\ast$Active methods used an earliest-first two-probe budget and never tested
hop 3, so these are censored cells rather than 0.00 accuracy.}
\label{tab:content-acc}
\end{table}

Depth 3 is reportable for the post-hoc methods: \dr{} remains 0.00 and the
judge is 0.57 [0.41,0.73]. On MuSiQue, the Claude-agent judge predicts hop 3
in 2/20, 4/18, and 18/25 failures at depths 1--3, so fixed late-position bias
does not explain its depth-3 score. It is the only non-degenerate method tested
at depth 3 because active cells are budget-censored. A fair depth-2 paired
comparison of \pa{} with the best post-hoc envelope has 11
\pa{}-only versus 6 envelope-only correct cases (19 both, 15 neither;
$n{=}51$), exact McNemar $p{=}0.33$. At depth 2 the envelope's correct set is
the judge's; there is no significant head-to-head difference.

The bridge-entity slice comprises failed scoring trajectories only, pooled
across all four content conditions and restricted to bridge entities: at depth
1, \sr{} is 0.89 and \pa{} is 0.70
($n{=}27$); at depth 2, \sr{} is 0.14 and \pa{} is 0.43 ($n{=}7$); no depth-3
bridge cases exist. The depth-2 slice is too small to support a mechanism claim
either way, and the depth-1 contrast cannot establish that suffix regeneration
is generally suited to bridge corruption.

\subsection{Answer Outcomes After Content Corruption}
\label{sec:outcomes-results}

Attribution evaluates only injected trajectories that remain failed. The
certified span pair also makes it possible to measure what happened to every
generated answer, including trajectories that recover. Table~\ref{tab:absorption}
reports absorbed, resisted, and derailed outcomes over all content-fault
interventions. The later-depth rows contain all injected cases at their
respective depths, not only the failed cases used in the attribution table.
Recovery is 0.71 and 0.70 pooled at depths 1--2, but 0.27 at depth 3; it is
not uniformly high. Depth 3 is majority derailed (0.56).

\begin{table}[t]
\centering
\small
\begin{tabular}{lrrrr}
\hline
Depth & $n$ & Absorbed & Resisted & Derailed \\
\hline
1 & 311 & 32 (0.10) & 220 (0.71) & 59 (0.19) \\
2 & 168 & 20 (0.12) & 117 (0.70) & 31 (0.18) \\
3 & 48 & 8 (0.17) & 13 (0.27) & 27 (0.56) \\
\hline
\end{tabular}
\caption{Deterministic outcomes over all injected content cases.}
\label{tab:absorption}
\end{table}

Absorption is strongly answer-shaped. Within type, answer facts are absorbed in
53/346 (0.15) interventions, bridge entities in 7/107 (0.07), and salient
entities in 0/74 (0.00); answer facts make up 53/60 absorbed outcomes.
Query contamination is 19/301 (0.06) among interventions with at least one
later query, the correct restricted denominator; it is 19/527 (0.04) over all
interventions. Thus later query text rarely contains the changed span even when
the answer outcome is affected.

\section{Discussion}
\label{sec:discussion}

\paragraph{Post-hoc signal after suffix re-execution.}
Coverage gating attributes a failure to the earliest hop with low answer
support. Because it scores gold-final-answer coverage, intermediate hops often
fail its test without a causal fault. Its deep zeros therefore do not by
themselves show suffix re-execution removed signal: it predicts hop 1 in 96\%
of structural failures.

The stronger observation is that the LLM judge and frozen-hop repair localize
only 0.25 to 0.48 of failures at hops 2 and 3. Whether regenerated suffixes
overwrite signal, rather than difficulty already being present in static edited
traces, is a hypothesis for paired static-versus-live and no-op controls.

This account does not require every later hop to be wrong. A later
retrieval can be topically relevant and still be a downstream consequence of a
faulty earlier premise. It can also partially repair an earlier failure without
restoring an identifiable trace signature. The relevant distinction is between
the information an agent needs to answer and the information a post-hoc
diagnoser needs to identify the intervention hop. The former can be recovered
by compensating evidence even when the latter has been lost.

\paragraph{Counterfactual repair under downstream dependence.}
Frozen-hop repair tests a useful causal question: does repairing candidate hop
$h$ change the answer when all later observed hops are held fixed? It can
mislocalize when those later hops were generated from the original fault.
Repairing the true early cause may not change the final answer because a later
hop still carries the corrupted dependency. Conversely, repairing a later
answer-bearing hop can restore correctness and receive credit despite being a
downstream consequence. This is why broad stage identification can be more
favorable than exact injected-hop attribution.

Suffix regeneration rebuilds downstream hops after repairing a candidate, while
frozen-hop repair preserves observed downstream evidence. Their content depth-2
contrast, 0.14 versus 0.59, is compatible with this difference in
counterfactual scope but does not establish a universal ranking. In the
bridge-entity depth-2 slice, \sr{} is 0.14 and \pa{} is 0.43 on $n{=}7$,
whereas the depth-1 slice is 0.89 versus 0.70 on $n{=}27$. Thus the depth-2
slice is too small to support a mechanism claim either way.

The two probes should therefore not be read as interchangeable versions of the
same repair. Frozen-hop repair asks whether the observed downstream trajectory
would support a correct answer after changing one local input. Suffix
regeneration asks whether a repaired prefix can lead the agent to a different
downstream trajectory. Each question is useful, but each can award credit to a
different position in a causal chain. Exact-hop evaluation exposes this
difference directly, whereas an answer-only evaluation would collapse both
repairs into a single correctness outcome.

Both active methods stop at the first restoring repair in earliest-first order,
creating a potential early-hop prior.
At content depth 3, both active methods predict hop 3 on 0/35 cases and an
earlier hop on 29/35, mechanically explained by the two-probe cap. The
structural \pa{} pattern is non-monotonic rather than simply early-biased.
Exhaustive and reverse-order ablations are enabled in the benchmark
harness\footnote{\url{https://github.com/anote-ai/Research-AgenticRAG}} and are
designated follow-up work, including a budget-4 depth-3 active re-diagnosis.

\paragraph{Recovery as a separate outcome.}
Live interventions reveal a second property that static trace edits cannot
measure: the agent can sometimes heal a corrupted trajectory through a later
retrieval. This is not a failure of the intervention. It identifies a case in
which the selected fault did not reach the final answer, so no intervention
site should be credited for an end-to-end failure. The distinction is particularly important
for content corruption, where the executed trajectory contains a changed span
but later retrieval can access clean corpus evidence. A persistent corpus-level
corruption would test a different mechanism and remains future work rather
than a feature of the present evaluation.

\paragraph{Implications for diagnosis.}
The appropriate repair probe depends on the path by which information flows
through the trajectory. A single universal localizer must choose between
preserving downstream evidence and regenerating it, and either choice can hide
a different fault family. Future path-aware or cascaded diagnosis can select a
probe based on suspected fault type, then report uncertainty when multiple hops
remain causally entangled. At content depth 2, the judge uses 1,941.24 tokens
per diagnosis and 3,960.12 per correct localization; \pa{} uses 3,382.35 and
5,750.00, while \sr{} uses 4,041.02 and 29,441.71. In the structural sweep at
depth 2, judge and \pa{} cost per correct localization is 9,062.11 and
18,107.78. Appendix~\ref{app:cost} reports all depths and methods. These are
oracle-assisted instrument costs, not online deployment costs.

\section{Limitations}

The dense Claude Haiku 4.5 MuSiQue sweep has adequate failed-case counts at all
three requested depths ($43$, $36$, and $21$). A matching dense
HotpotQA sweep has only 9 and 2 failed trajectories at its valid depths because
the agent recovers nearly all injected faults; it is therefore descriptive and
excluded from the main comparison. The paper does not claim a complete
backbone-by-dataset factorial evaluation, and it excludes CRAG from headline
results because CRAG is normalized here as a single-turn benchmark.

The content intervention corrupts a trajectory copy while the retrieval corpus
remains clean, allowing re-retrieval to heal some faults. Pooled recovery is
0.71, 0.70, and 0.27 at depths 1--3. The coverage baseline is an adaptation
that scores final-answer coverage, and a hop-specific coverage target could
behave differently. The persisted content trajectories do not retain aligned
hop-level injected suffixes, so sub-query-change and document-overlap rates
cannot be computed post hoc. The benchmark harness now persists these details
and supports paired static-versus-live and
suffix-divergence analyses; these are designated follow-up measurements.
Exact-hop accuracy is intentionally strict, and stage-level scoring can credit
a diagnoser that finds the failure type but not its injected hop. LLM judges
are sensitive to prompting and token cost. In the structural sweep, the LLM
judge's apparent depth robustness can include a positional prior.

Providers ran at their default sampling temperature. A no-op resumed-suffix
control is therefore required to bound spontaneous failure rather than assume
determinism; it is implemented in the benchmark harness and remains follow-up
work. The same harness supports earliest, reverse, and exhaustive probe orders
with per-probe outcomes. The content depth-3 active cells are censored by the
two-probe budget, not evidence that active localization fails at hop 3; a
budget-4 depth-3 re-diagnosis is required.

\section{Conclusion}

\ours{} evaluates causal failure attribution by injecting a known fault and
allowing the agent to generate its downstream response. In the dense
Claude-MuSiQue sweep, coverage-based post-hoc attribution is 0.91 at hop
1 and 0.00 at hops 2 and 3, with 43, 36, and 21 failed trajectories
respectively; it predicts hop 1 in 96\% of failures, so it is an earliest-hop
rule, and no evaluated diagnoser exceeds 0.48 exact-hop accuracy at structural
hops 2 and 3. In content runs at depth 2, coverage, LLM judge, frozen-hop, and
suffix-regeneration accuracy are 0.00, 0.49, 0.59, and 0.14 ($n{=}51$); the
judge and frozen-hop repair do not significantly differ (exact McNemar
$p{=}0.33$). This evidence distinguishes coverage's earliest-hop behavior from
the two scopes of oracle-assisted counterfactual repair. Future work should
execute no-op, probe-order, budget-4, and suffix-divergence follow-ups before
generalizing beyond these realized failure populations.

\bibliography{references}

\appendix

\section{Reproducibility details}
\label{app:repro}

\paragraph{Retrieval and corpora.}
Dense retrieval uses \texttt{sentence-transformers/}\allowbreak
\texttt{all-MiniLM-L6-v2}:
normalized query and document embeddings are ranked by their dot product,
hence cosine similarity. The corpus is constructed separately for each
question. HotpotQA flattens each context title and its sentences into one
document; MuSiQue formats each paragraph as its title followed by paragraph
text. The content comparisons use BM25, whereas the structural headline is
the dense condition.

\paragraph{LLM-judge prompt.}
The system prompt is quoted verbatim:
\begin{quote}\small
You are a diagnostic judge for a multi-hop retrieval agent. Given the question,
the per-hop retrieved evidence, the final answer, and the gold answer, identify
the earliest hop where the agent went wrong. Respond with a single JSON object:
\{"stage": "retrieval"|"answer\_generation"|"none", "hop":
<1-based hop number, or 0 if none>\}.
\end{quote}
The user-prompt template is:
\begin{quote}\small\ttfamily\raggedright
QUESTION: \{trace.query\}\par
\par
HOPS:\par
\quad Hop \{i + 1\} query: \{q\}\par
\quad\quad - \{d\}\par
\quad\quad - (no documents retrieved)\par
\par
FINAL ANSWER: \{trace.final\_answer\}\par
GOLD ANSWER: \{gold\}
\end{quote}
The HOPS block is emitted for each query/evidence pair; its document line is
emitted once per document, and the no-documents line is emitted only for an
empty document list.

\paragraph{Probe budgets.}
The structural setting uses an earliest-first four-probe budget. The content
setting uses a two-probe budget.

\section{Per-condition content localization}
\label{app:content-by-condition}

Table~\ref{tab:content-by-condition} gives the content breakdown. A
dagger marks a failed denominator below 10. Active depth-3 entries are censored
because the budget prevents a hop-3 repair test.
\begin{table*}[t]
\centering\small
\begin{tabular}{llccc}
\hline
Condition & Diagnoser & Depth 1 & Depth 2 & Depth 3 \\
\hline
Claude--MuSiQue & \dr{} & 1.00 (20) & 0.00 (18) & 0.00 (25) \\
& \lj{} & 0.15 (20) & 0.33 (18) & 0.72 (25) \\
& \pa{} & 0.70 (20) & 0.39 (18) & --$^\ast$ \\
& \sr{} & 0.80 (20) & 0.22 (18) & --$^\ast$ \\
\hline
Claude--HotpotQA & \dr{} & 1.00 (23) & 0.00 (17) & -- \\
& \lj{} & 0.43 (23) & 0.71 (17) & -- \\
& \pa{} & 0.96 (23) & 0.76 (17) & -- \\
& \sr{} & 1.00 (23) & 0.06 (17) & -- \\
\hline
GPT-4o-mini--MuSiQue & \dr{} & 1.00 (11) & 0.00 (10) & 0.00 (10) \\
& \lj{} & 0.09 (11) & 0.40 (10) & 0.20 (10) \\
& \pa{} & 0.73 (11) & 0.40 (10) & --$^\ast$ \\
& \sr{} & 0.91 (11) & 0.20 (10) & --$^\ast$ \\
\hline
GPT-4o-mini--HotpotQA & \dr{} & 1.00 (37) & 0.00 (6)$^\dagger$ & -- \\
& \lj{} & 0.68 (37) & 0.50 (6)$^\dagger$ & -- \\
& \pa{} & 0.97 (37) & 1.00 (6)$^\dagger$ & -- \\
& \sr{} & 0.95 (37) & 0.00 (6)$^\dagger$ & -- \\
\hline
\end{tabular}
\caption{Exact-hop content localization. Parentheses give failed-case $n$.
$^\ast$Budget-censored active depth-3 cell, not a zero estimate.}
\label{tab:content-by-condition}
\end{table*}

\section{Additional content-fault outcome details}
\label{app:outcomes}

The deterministic outcome counts are in Table~\ref{tab:absorption}. Across all
depths, answer facts are absorbed in 53/346 (0.15) interventions, bridge
entities in 7/107 (0.07), and salient entities in 0/74 (0.00); answer facts
make up 53/60 absorbed outcomes. Query contamination is
19/301 (0.06) only among interventions with a later query and 19/527 (0.04)
over all interventions. The fair depth-2 paired comparison of \pa{} with the
post-hoc envelope has 11 \pa{}-only and 6 envelope-only cases (19 both, 15
neither; $n{=}51$; McNemar $p{=}0.33$).

\begin{figure*}[t]
\centering
\includegraphics[width=0.48\textwidth]{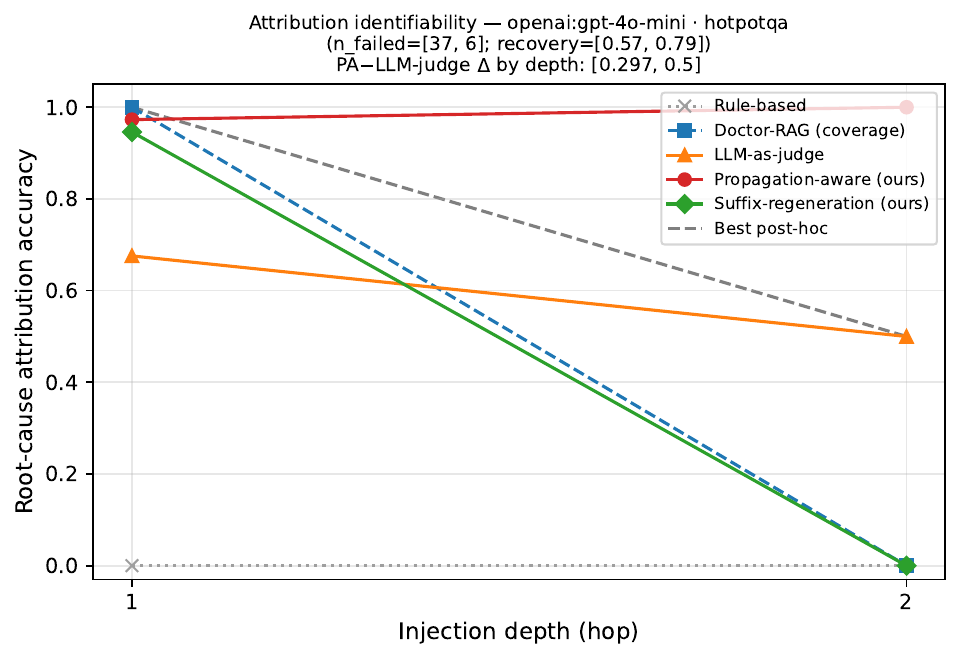}\hfill
\includegraphics[width=0.48\textwidth]{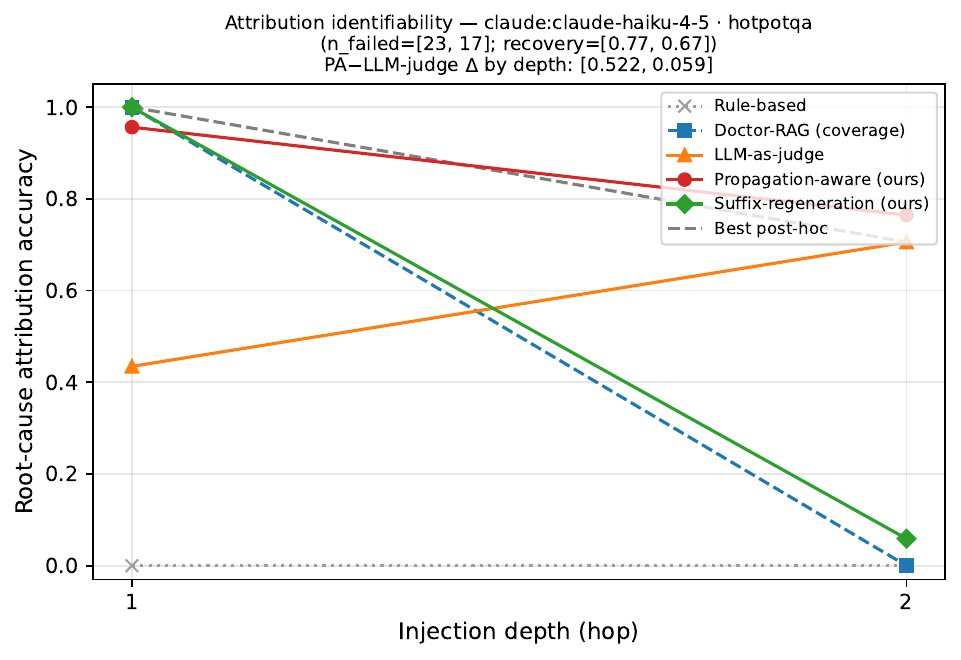}
\caption{Content-corruption curves from runs on HotpotQA with BM25 at
depths 1--2 by design. Left: GPT-4o-mini agent judged by Claude Haiku 4.5. Right: Claude
Haiku 4.5 agent judged by GPT-4o-mini. The panel annotation ``PA--LLM-judge
$\Delta$ by depth'' is the per-depth difference between Propagation-Aware and
LLM-Judge exact-hop accuracy. The dashed best post-hoc envelope line is the
element-wise maximum of the rule-based, coverage, and LLM-Judge accuracies.}
\label{fig:corruption-curves}
\end{figure*}

Figure~\ref{fig:corruption-curves} shows the HotpotQA BM25
content-corruption curves at depths 1--2 by design.

\section{Worked intervention example}
\label{app:example}

\noindent\fbox{\begin{minipage}{0.96\linewidth}\small
\textbf{Claude--MuSiQue, certified hop 2 bridge entity.} The question asks who
is the president of the new country that jointly established a Commission on
Truth and Friendship with the birth country of Satyo Husodo. The gold answer is
Francisco Guterres. The base trace asks Satyo Husodo birth country, establishing
Indonesia; Commission on Truth and Friendship Indonesia, establishing East
Timor; and president of Timor Leste East Timor, retrieving Francisco Guterres.
The certified span pair at hop 2 is \emph{East Timor $\rightarrow$ Chris
Harrison}. The resumed suffix uses three iterations and says the president's
name is not provided, so the outcome is derailed and query contamination is
false. The calls are \rb{}: answer generation, hop 0; \dr{}: retrieval, hop 1;
\lj{}: retrieval, hop 3; \pa{}: retrieval, hop 2; and \sr{}: retrieval, hop 1.
Only \pa{} matches the certified injection. Persisted records establish the
calls but not probe rationales or aligned injected-suffix evidence.
\end{minipage}}

\section{Cost per correct localization}
\label{app:cost}

Table~\ref{tab:correct-counts} gives correct localizations on failed
trajectories. Table~\ref{tab:cost} gives stored diagnosis-token costs, and
Table~\ref{tab:structural-ci} compares per-trajectory and clustered intervals.
Rule and \dr{} use zero diagnosis tokens by construction.
\begin{table*}[t]
\centering\small
\begin{tabular}{llrrrrr}
\hline
& & \multicolumn{5}{c}{Correct / $n$} \\
Population & Depth & \rb{} & \dr{} & \lj{} & \pa{} & \sr{} \\
\hline
Structural & 1 & 1/43 & 39/43 & 11/43 & 22/43 & -- \\
Structural & 2 & 0/36 & 0/36 & 9/36 & 9/36 & -- \\
Structural & 3 & 0/21 & 0/21 & 9/21 & 10/21 & -- \\
\hline
Content & 1 & 0/91 & 91/91 & 39/91 & 80/91 & 84/91 \\
Content & 2 & 0/51 & 0/51 & 25/51 & 30/51 & 7/51 \\
Content & 3 & 0/35 & 0/35 & 20/35 & 0/35 & 0/35 \\
\hline
\end{tabular}
\caption{Correct exact-hop localizations among failed trajectories. A dash
means the diagnoser was not evaluated in that cell.}
\label{tab:correct-counts}
\end{table*}

\begin{table*}[t]
\centering\small
\begin{tabular}{llrrr}
\hline
Population & Depth & \lj{} & \pa{} & \sr{} \\
& & \multicolumn{3}{c}{Tokens/correct \quad / \quad Tokens/diagnosis} \\
\hline
Structural & 1 & 8,840.18 / 2,261.44 & 9,461.41 / 4,840.72 & -- \\
Structural & 2 & 9,062.11 / 2,265.53 & 18,107.78 / 4,526.94 & -- \\
Structural & 3 & 6,154.22 / 2,637.52 & 9,452.20 / 4,501.05 & -- \\
\hline
Content & 1 & 3,470.69 / 1,487.44 & 2,422.49 / 2,129.66 & 2,845.94 / 2,627.02 \\
Content & 2 & 3,960.12 / 1,941.24 & 5,750.00 / 3,382.35 & 29,441.71 / 4,041.02 \\
Content & 3 & 4,664.45 / 2,665.40 & -- & -- \\
\hline
\end{tabular}
\caption{Stored diagnosis-token cost. A dash means no finite cost or an
untested cell; content active depth-3 cells are budget-censored.}
\label{tab:cost}
\end{table*}

\begin{table*}[t]
\centering\small
\begin{tabular}{llcc}
\hline
Diagnoser & Depth & Per-trajectory CI & Clustered CI \\
\hline
\dr{} & 1/2/3 & [0.81,0.98]/[0,0]/[0,0] & [0.80,0.98]/[0,0]/[0,0] \\
\lj{} & 1/2/3 & [0.12,0.40]/[0.11,0.42]/[0.24,0.67] & [0.10,0.42]/[0.13,0.38]/[0.19,0.71] \\
\pa{} & 1/2/3 & [0.37,0.67]/[0.11,0.39]/[0.29,0.67] & [0.36,0.67]/[0.11,0.38]/[0.24,0.75] \\
\hline
\end{tabular}
\caption{Structural CI comparison, in depth order 1/2/3. Main tables use
base-question-clustered intervals.}
\label{tab:structural-ci}
\end{table*}

\section{Substrate recovery and predicted-hop checks}
\label{app:substrates}

Each Table~\ref{tab:recovery-substrates} cell is total interventions / failed
interventions / recovery. Near-total HotpotQA recovery leaves too few failures
for a headline depth comparison. FRAMES offers a different substrate but not
the dense headline; CRAG is excluded because its normalized
setting is single-turn. These are informative negatives for substrate choice.
Table~\ref{tab:structural-predictions} reports the structural predicted-hop
checks.
\makeatletter
\setlength{\@dblfptop}{0pt}
\makeatother
\begin{table*}[t]
\centering\small
\begin{tabular}{llll}
\hline
Condition and role & Hop 1 & Hop 2 & Hop 3 \\
\hline
Claude--FRAMES, BM25 substrate check & 52 / 18 / 0.65 & 52 / 19 / 0.63 & 36 / 17 / 0.53 \\
Claude--HotpotQA, dense recovery negative & 248 / 9 / 0.96 & 128 / 2 / 0.98 & -- \\
Claude--MuSiQue, dense structural headline & 240 / 43 / 0.82 & 224 / 36 / 0.84 & 120 / 21 / 0.82 \\
Content pooled, BM25 & 311 / 91 / 0.71 & 168 / 51 / 0.70 & 48 / 35 / 0.27 \\
CRAG, normalized single-turn exclusion & not a depth sweep & not a depth sweep & not a depth sweep \\
\hline
\end{tabular}
\caption{Recovery across selected substrates and the content
pool.}
\label{tab:recovery-substrates}
\vspace{1.5ex}
\begin{tabular}{lrrr}
\hline
Diagnoser & Depth 1 & Depth 2 & Depth 3 \\
\hline
\dr{} predicts hop 1 & 39/43 (0.91) & 36/36 (1.00) & 21/21 (1.00) \\
\lj{} predicts hop 3 & 14/43 (0.33) & 12/36 (0.33) & 9/21 (0.43) \\
\pa{} exact-hop accuracy & 0.51 & 0.25 & 0.48 \\
\hline
\end{tabular}
\caption{Structural predicted-hop checks. \dr{} is an empirical earliest-hop
rule in the deep failed sets; the judge has a late-hop tendency.}
\label{tab:structural-predictions}
\end{table*}

\end{document}